\documentclass{llncs}
\usepackage{latexsym}
\usepackage{CJK}
\usepackage{graphicx}
\usepackage{multirow}
\usepackage{url}
\usepackage{makeidx}  
\begin{document}
\frontmatter          
\pagestyle{headings}  

\title{Transfer Deep Learning for \\Low-Resource Chinese Word Segmentation \\with a Novel Neural Network}
\titlerunning{Hamiltonian Mechanics}  
%
\author{Jingjing Xu\inst{1,2}  \and Shuming Ma \inst{1,2}  \and Yi Zhang\inst{1,2}  \and Bingzhen Wei\inst{1,2} \and  \\ Xiaoyan Cai\inst{3} \and Xu Sun\inst{1,2}  }
\authorrunning{Ivar Ekeland et al.} 
%
\tocauthor{Ivar Ekeland, Roger Temam, Jeffrey Dean, David Grove,
Craig Chambers, Kim B. Bruce, and Elisa Bertino}
\institute{ MOE Key Laboratory of Computational
Linguistics, Peking University,\\ Beijing, China 
\and
School of Electronics Engineering and Computer Science,
Peking University,\\ Beijing, China \\
\and
School of Automation, Northwestern Polytechnical University,\\ Xi'an, China\\
\email{jingjingxu@pku.edu.cn}}

\maketitle              

\begin{abstract}
Recent studies have shown effectiveness in using neural networks for Chinese word segmentation. However, these models rely on large-scale data and are less effective for low-resource datasets because of insufficient training data. We propose a transfer learning method to improve low-resource word segmentation by leveraging high-resource corpora. First, we train a teacher model on high-resource corpora and then use the  learned knowledge to initialize a student model. Second, a weighted data similarity method is proposed to train the student model on low-resource data. Experiment results show that our work significantly improves the performance on low-resource datasets: 2.3\% and 1.5\% F-score on PKU and CTB datasets. Furthermore, this paper achieves  state-of-the-art results: 96.1\%, and 96.2\% F-score on PKU and CTB datasets~\footnote{http://jingjingxu.com/tutorial.pdf}. 
\end{abstract}
\section{Introduction}

Chinese word segmentation (CWS) is an important step in Chinese natural language processing. The most widely used approaches~\cite{Peng,Xue03} treat CWS as a sequence labelling problem in which each character is assigned with a tag. Formally, given an input sequence $\mathbf{x}=x_{1}x_{2}...x_{n}$, it produces a tag sequence $\mathbf{y}=y_{1}y_{2}...y_{n}$. Many exsiting techniques, such as conditional random fields, have been successfully applied to CWS ~\cite{Lafferty,Sun2011Enhancing,SunLWL14,Tseng05,ZhaoHLL10}. However, these approaches incorporate many handcrafted features. Therefore, the generalization ability is restricted.

In recent years, neural networks have become increasingly popular in CWS, which focused more on the ability of automated feature extraction. Collobert et al.~\cite{Collobert} developed a general neural architecture for sequence labelling tasks. Pei et al.~\cite{pei} used convolutioanl neural networks to capture local features within a fixed size window. Chen et al.~\cite{chen} proposed gated recursive neural networks to model feature combinations. The gating mechanism was also used by Cai and Zhao~\cite{Cai2016Neural}.

%
%
%
%
%
%
%
%
%
%

However, this success relies on massive labelled data and are less effective on low-resource datasets. The major problem is that a small amount of labelled data leads to inadequate training and negatively impacts the ability of generalization. However, there are enough corpora which consist of massive annotated texts. All can be used to improve the task. Thus, we propose a transfer learning method to address the problem by leveraging high-resource datasets. 

First, we train a teacher model on high-resource datasets and then use the learned knowledge to initialize a student model.  Previous neural network models usually use random initialization which relies on massive labelled data. It is hard for a randomly initialized model to achieve the expected results on low-resource datasets. Motivated by that, we propose a teacher-student framework to initialize the student model.
However, it is hard to directly make use of high-resource datasets to train the student model because different corpora have different data distributions. The shift of data distributions is a major problem. To address the problem, we propose a weighted data similarity method which computes a similarity of each high-resource sample with a low-resource dataset. Experiment results show that using our transfer learning method, we substantially improve the performance on low-resource datasets.





\begin{CJK*}{GBK}{song}

 

%

\end{CJK*}

With the increasing of layers  which are designed to improve the ability of feature extraction, the training speed is becoming limit. To speed up training, we explore mini-batch asynchronous parallel (MAP) learning on neural segmentation in this paper. Existing asynchronous parallel learning methods are mainly for sparse models~\cite{echtRWN11}. For dense models, like neural networks, asynchronous parallel methods bring inevitable gradient noises. However, the theoretical analysis  by Sun~\cite{Sun2016Asynchronous} showed that the learning process with gradient errors can still be convergent on neural models. Motivated by that, we explore the MAP approach on neural segmentation in this paper. The parallel method accelerates training substantially and the training speed is almost five times faster than a serial mode.

The main contributions of the paper are listed as follows:
\begin{itemize}

\item A transfer learning method is proposed to improve low-resource word segmentation by leveraging high-resource corpora. 


\item To speed up training, mini-batch asynchronous parallel learning on neural word segmentation is explored. 

\end{itemize}

\section{Transfer Learning by Leveraging High-Resource Datasets}

Previous neural word segmentation models are less effective on low-resource datasets since these models only focus on in-domain supervised learning. Furthermore, there are enough corpora which consist of massive annotated texts. For scenarios where we have insufficient labelled data, transfer learning is an effective way to improve the task. Motivated by that, we propose a transfer learning method to leverage high-resource corpora.

 First, we propose a teacher-student framework to initialize a model with the learned knowledge. We train a teacher model on a dataset where there is a large amount of training data (e.g., MSR). The learned parameters are used to initialize a student model. Therefore, the student model is trained from the learned parameters, rather than randomly initialization.
 
  Second, the student model is trained by the weighted data similarity method. However, since different corpora have different data distributions, it is hard to directly make use of high-resource datasets to train the student model. Thus, to avoid the shift of data distributions, high-resource corpora are used to train the student model based on the weighted data similarity method. This method identifies the similarity of each high-resource sample with a low-resource dataset. We use different learning rates for different samples. A learning rate is adjusted by the weighted data similarity automatically. The weighted data similarity $w^{t}_{i}$ is updated as follows. 
 
 %

First, calculate the update rate $a^{t}$:
\begin{eqnarray}
e^{t}=(1-\frac{2*p^{t}*r^{t}}{p^{t}+r^{t}})
\end{eqnarray}
\begin{eqnarray}
{a}^{t}=\frac{1}{2}log{\frac{1-e^{t}}{e^{t}}}
\end{eqnarray}
where $p^{t}$ and $r^{t}$ are precision and recall of the student model on high-resource data. The update rate $a^{t}$ is determined by the error rate $e^{t}$. The error rate is a simple and effective way to evaluate the data similarity. 

Next, update the data similarity  after $t$ iterations: 
\begin{eqnarray}
S^{t+1}=(w^{t+1}_{1},{\ldots},w^{t+1}_{i},{\ldots},w^{t+1}_{N})
\end{eqnarray}
\begin{eqnarray}
w^{t+1}_{i}=\frac{w^{t}_{i}}{Z^{t}*m}\sum_{j=1}^{m}{exp({a}^{t}I(y_{i,j}=p_{i,j}))}
\end{eqnarray}
where $m$ is the length of sample $i$, $I()$ is the indicator function which evaluates whether the prediction $p_{i,j}$ is equal with the gold label $y_{i,j}$ or not and $Z^{t}$ is the regularization factor which is computed as:
\begin{eqnarray}
Z^{t}=\sum_{i}{\frac{w^{t}_{i}}{m}\sum_{j=1}^{m}{exp({a}^{t}I(y_{i,j}=p_{i,j}))}}
\end{eqnarray}

Finally, the weighted data similarity is used to compute the learning rate $\alpha^{t}_{i}$:
\begin{eqnarray}
\alpha^{t}_{i}=\alpha^{t}*w^{t}_{i}
\end{eqnarray}
where $\alpha^{t}$ is the fixed learning rate for a low-resource dataset,
$w^{t}_{i}$ indicates the similarity between sentence $i$
and a low-resource corpus, which ranges from 0 to 1.
\section{Unified Global-Local Neural Networks}

Insufficient data puts forward higher requirements for feature extraction. Our key idea is to combine several kinds of weak features to achieve the better performance. Unlike previous networks which focus on a single kind of feature: either complicated local features or global dependencies, our network has an advantage of combining complicated local features with long dependencies together. Both of them are necessary for CWS and should not be neglected. Our network is built on a simple encoder-decoder structure. A encoder is designed to model local combinations and a decoder is used to capture long distance dependencies.

First, words are represented by embeddings stored in a lookup table $D^{|v|*d}$ where $v$ is the number of words in the vocabulary and $d$ is the embedding size. The lookup table is pre-trained on giga-word corpus where unknown words are mapped to a special symbol. The inputs to our model are $x_{1},x_{2},...,x_{n}$ which are represented by $D={D_{x_{1}},D_{x_{2},...,D_{x_{n}}}}$.

We first extract a window context $H^{0}{\in}R^{n,k,d}$ from an input sequence which is padded with special symbols according to the window size:
\begin{eqnarray}
H^{0}_{i,j}=D[i+j]
\end{eqnarray}
where $n$ is the sentence length, $k$ is the window size and $d$ is the embedding length. $H^{0}$ will be input to the encoder to produce complicated local feature representations.

\textbf{Encoder.} The encoder is composed of filter recursive networks. According to the filter size, we first choose every patch and input it to gate function to get next layer $H^1{\in}R^{n,k-f_1+1,d}$ where $f_1$ is the filter size of $1^{th}$ hidden layer.

In a gate cell of filter recursive networks, output $H^1$ of the $i^{th},j^{th}$ hidden node is computed as:
\begin{eqnarray}
H^{1}_{i,j}=z_{h}{\odot}h^{'}+\sum_{d_{i}=0}^{f_1-1}(z_{d_{i}}{\odot}H^{0}_{i,j+d_{i}})
\end{eqnarray}
where $z_{h}$ and $z_{d_{i}}$ are update gates for new activation $h^{'}$ and inputs, while ${\odot}$ means element-wise multiplication.
To simplify the cell, $z_{h}$, $z_{d_{i}}$  are computed as:
\begin{eqnarray}
\left[
\begin{array}{c}
z_{h}  \\
z_{0} \\
...\\
z_{d_{i}}\\
...\\
z_{f_1-1}
\\
\end{array}
\right]
=sigmoid(U
\left[
\begin{array}{c}
 h^{'}\\
 H^{0}_{i,j}\\
\ldots\\
 H^{0}_{i,j+d_{i}}\\
\ldots\\
  H^{0}_{i,j+f_1-1}\\
\end{array}
\right]
 )
\end{eqnarray}
where $U{\in}R_{(f_{1}+1)d*(f_{1}+1)d}$ and
the new activation $h^{'}$ is computed as:
\begin{eqnarray}
h^{'}
=tanh(W
\left[
\begin{array}{c}
 r_{0}{\odot}H^{0}_{i,j}\\
\ldots\\
 r_{d_{i}}{\odot}H^{0}_{i,j+d_{i}}\\
\ldots\\
 r_{f_{1}-1}{\odot}H^{0}_{i,j+f_1-1}
\end{array}
\right]
 )
\end{eqnarray}
where $W{\in}R_{d*f_{1}d}$ and $r_{0},...,r_{f_{1}-1}$ are reset gates for inputs, which can be formalized as:
\begin{eqnarray}
\left[
\begin{array}{c}
r_{0} \\
\ldots\\
r_{d_{i}}\\
\ldots\\
r_{f_{1}-1}
\end{array}
\right]
=sigmoid(G
\left[
\begin{array}{c}
H^{0}_{i,j}\\
\ldots\\
H^{0}_{i,j+d_{i}}\\
\ldots\\
H^{0}_{i,j+f_{1}-1}
\end{array}
\right]
 )
\end{eqnarray}

These operations will repeat until we get $H^{l}{\in}R^{n, 1,d}$ which is reduced dimension to $H^{l}{\in}R^{n,d}$.

\textbf{Decoder.} The decoder is composed of bi-directional long shor-term memory network (Bi-LSTM). The local features encoded by filter recursive neural networks are refined into global dependencies and then decoded to tag sequences in this stage.

\section{Mini-Batch Asynchronous Parallel Learning}


With the development of multicore computers, there is a growing interest in parallel techniques. Researchers have proposed several schemes~\cite{minibatch}, but most of them require locking so the speedup is limited. Asynchronous parallel learning methods without locking can maximize the speedup ratio. However, existing asynchronous parallel learning methods are mainly for sparse models. For dense models, like neural networks, asynchronous parallel learning brings gradient noises which are very common and inevitable. Read-read conflicts break the sequentiality of training procedure, read-write and write-write conflicts lead to incorrect gradients. Nevertheless, Sun~\cite{Sun2016Asynchronous} proved that the learning process with gradient errors can still be convergent. Motivated by that, we train our model in the  asynchronous parallel way.

We find that Adam~\cite{Kingma2014Adam} is a practical method to train large neural networks. Therefore, we run the asynchronous parallel method based on Adam training algorithm.


The training algorithm is realized without any locking. For each mini-batch, we uniformly distribute it into different processors. Processors compute the increment of gradient $\Delta w_{t}$ in parallel, where $w_{t}$ is stored in a shared memory and each processor can read and update it.

\begin{table*}
\caption{Comparisons between UGL and baselines on low-resource datasets: PKU and CTB. }
\centering
\begin{tabular}{|l|c|c|c|c|c|c|}
\hline
\multicolumn{1}{|c|}{\multirow{2}{1.2cm}{Models}}&\multicolumn{3}{|c|}{PKU}&\multicolumn{3}{|c|}{CTB}\\
\cline{2-7}
\multicolumn{1}{|c|}{}
&{P}&{R}&{F}&{P}&{R}&{F} \\
\hline

Bi-LSTM&94.1&92.6&93.3&94.2&94.5&94.3 \\
GRNN&94.5&93.6&94.0&94.8&94.9&94.8 \\
\hline
\textbf{UGL}&95.2&94.1&\textbf{94.6}&95.4&95.2&\textbf{95.3} \\ \hline
\end{tabular}

\label{tab2}
\end{table*}

\section{Experiments}
The proposed model is evaluated on three datasets: MSR, PKU and CTB. We treat MSR as a high-resource dataset, PKU and CTB as low-resurce datasets. MSR and PKU are provided by the second International Chinese Word Segmentation Bakeoff~\cite{Emerson}. CTB is from Chinese TreeBank 8.0 and split to training and testing sets in this paper. 
We randomly split 10\%
of the training sets to development sets which are used to choose the suitable hyper-parameters. All idioms, numbers and continuous English characters are replaced to special flags. The improvements achieved by an idiom dictionary are very limited, less than 0.1\% F-score on all datasets. The character embeddings are pretrained on Chinese gigaword by word2vec. All results are evaluated by $F_{1}$-score which is computed
by the standard Bakeoff scoring program. 

\begin{table}
\caption{Improvements of our proposal on low-resource datasets: PKU and CTB.
}
\centering
\begin{tabular}{|c|c|c|}
\hline
\multicolumn{1}{|c|}{\multirow{1}{*}{Models}}&\multicolumn{1}{|c|}{PKU}&\multicolumn{1}{|c|}{CTB}\\

\hline
Bi-LSTM&93.3&94.3\\

\hline
UGL&94.6&95.3\\

+Transfer Learning&\textbf{95.6}&\textbf{95.8}\\
\hline
\multicolumn{1}{|c|}{\textbf{Improvement}}&\textbf{2.3}&\textbf{1.5}\\

\multicolumn{1}{|c|}{\textbf{Error Rate Reduction}}&\textbf{34.3}&\textbf{26.3}\\
\hline
\end{tabular}
\label{transferresults}
\end{table}

\subsection{Setup}

Hyper-parameters are set according to the performance on development sets. We evaluate the mini-batch size $m$ in a serial mode and choose $m=16$. Similarly, the window size $w$ is set as 5, the fixed learning rate $\alpha$ is set as 0.01, the dimension of character embeddings and hidden layers $d$ is set as 100. $d=100$ is a good balance between model speed and performance. 

Inspired by Pei et al.~\cite{pei}, bigram features are applied to our model as well. Specifically, each bigram embedding is represented as a single vector. Bigram embeddings are initialized randomly. We ignore lots of bigram features which only appear once or twice since these bigram features not only are useless, but also make a bigram lookup table huge.

All experiments are performed on a commodity 64-bit Dell Precision T5810 workstation with one 3.0GHz 16-core CPU and 64GB RAM. The C\# multiprocessing module is used in this paper.
\begin{table*}[!htb]
\caption{Comparisons with state-of-the-art neural networks on lower-resource datasets: PKU and CTB.
}
\centering

\begin{tabular}{|l|c|c|c|c|c|c|c|}
\hline
\multicolumn{1}{|c|}{\multirow{2}{*}{}}&\multicolumn{1}{|c|}{\multirow{2}{*}{Models}}&\multicolumn{3}{|c|}{PKU}&\multicolumn{3}{|c|}{CTB}\\
\cline{3-8}
&\multicolumn{1}{|c|}{}
&{P}&{R}&{F}&{P}&{R}&{F} \\
\hline
\multicolumn{1}{|c|}{\multirow{4}{*}{Unigram}}&
Zheng et al.~\cite{ZhengCX13}&92.8& 92.0& 92.4&*&*&*\\

&Pei et al. \cite{pei}&94.4&93.6&94.0&*&*&*\\

&Cai and Zhao \cite{Cai2016Neural}&95.8&95.2&95.5&*&*&*\\

&\textbf{Our Work}&96.0&95.1&\textbf{95.6}&95.9&95.8&\textbf{95.8} \\
\hline
\multicolumn{1}{|c|}{\multirow{4}{*}{Bigram}}&
Pei et al. \cite{pei}&*&*&95.2&*&*&*\\
&Ma and Hinrichs \cite{ma}&*&*&95.1&*&*&*\\
&Zhang et al. \cite{zhang2016transition}&*&*&95.7&*&*&*\\
&\textbf{Our Work}&96.3&95.9&\textbf{96.1}&96.2&96.1&\textbf{96.2}\\

\hline


\end{tabular}
\label{tabneural}
\end{table*}

\begin{table*}[!htb]
\caption{Comparisons with previous traditional models on lower-resource datasets: PKU and CTB.
}
\centering

\begin{tabular}{|l|c|c|c|c|c|c|}
\hline
\multicolumn{1}{|c|}{\multirow{2}{*}{Models}}&\multicolumn{3}{|c|}{PKU}&\multicolumn{3}{|c|}{CTB}\\
\cline{2-7}
\multicolumn{1}{|c|}{}
&{P}&{R}&{F}&{P}&{R}&{F} \\
\hline
Tseng et al. \cite{Tseng05}&*&*&95.0&*&*&*\\
Zhang et al. \cite{Zhang2006}&*&*&95.1&*&*&*\\
Zhang and Clark ~\cite{zhangclark2007}&*&*&94.5&*&*&*\\
Sun et al. \cite{sunxu}&*&*&95.4&*&*&*\\
\hline
\textbf{Our Work}&96.3&95.9&\textbf{96.1}&96.2&96.1&\textbf{96.2}\\
\hline

\end{tabular}
\label{tab3}
\end{table*}

\begin{figure}[!hbt]
\centerline{\includegraphics[width=5cm]{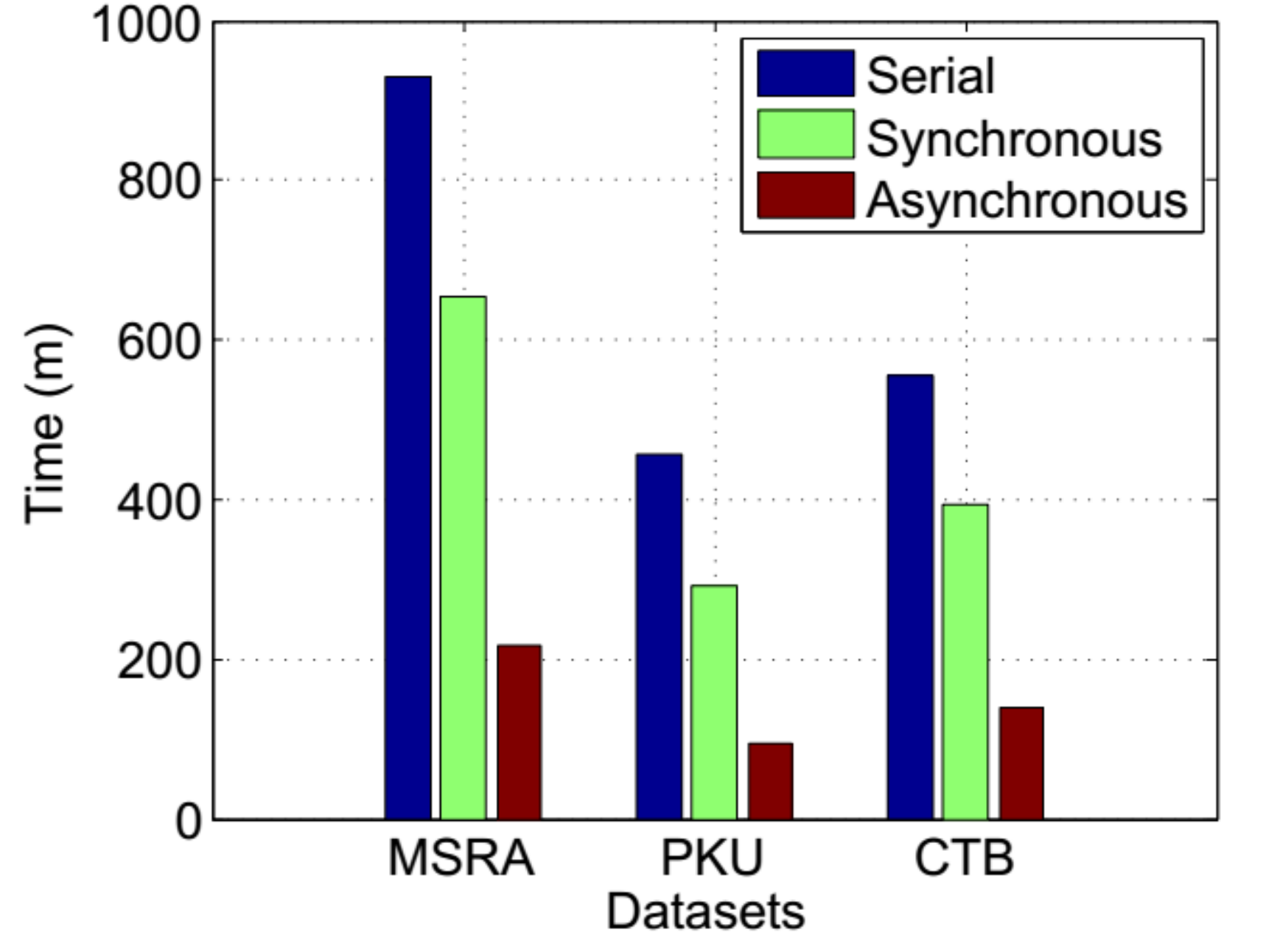}}
\caption{Comparisons of training time among serial, synchronous and asynchronous algorithms on three datasets.}
\label{parallelfscoreq}
\end{figure}

\subsection{Results and Discussions}
\begin{figure}[!hbt]

\begin{minipage}{0.33\linewidth}
\centerline{\includegraphics[width=4.1cm]{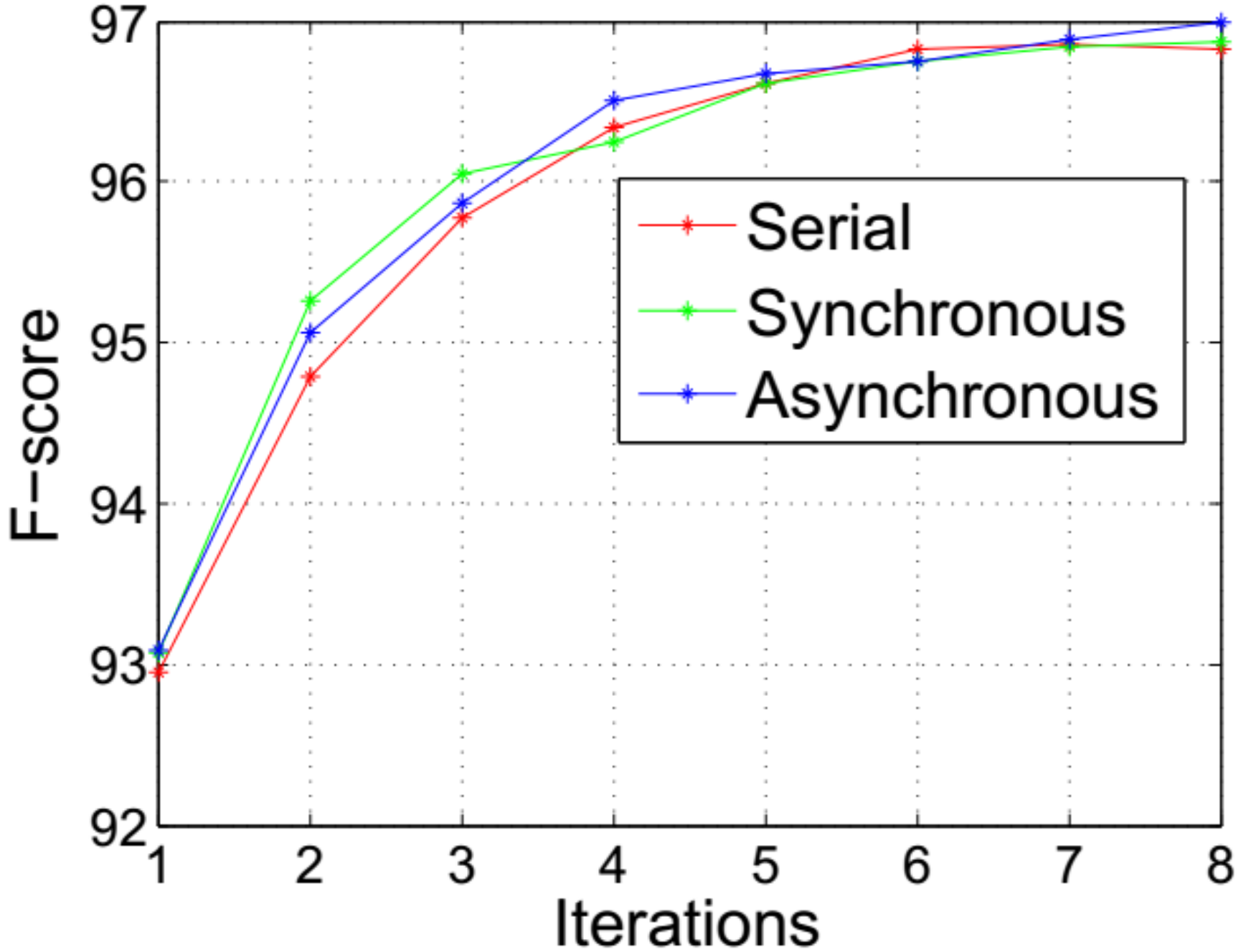}}
\centerline{(a) MSR}
\end{minipage}
\begin{minipage}{0.32\linewidth}
\centerline{\includegraphics[width=4.1cm]{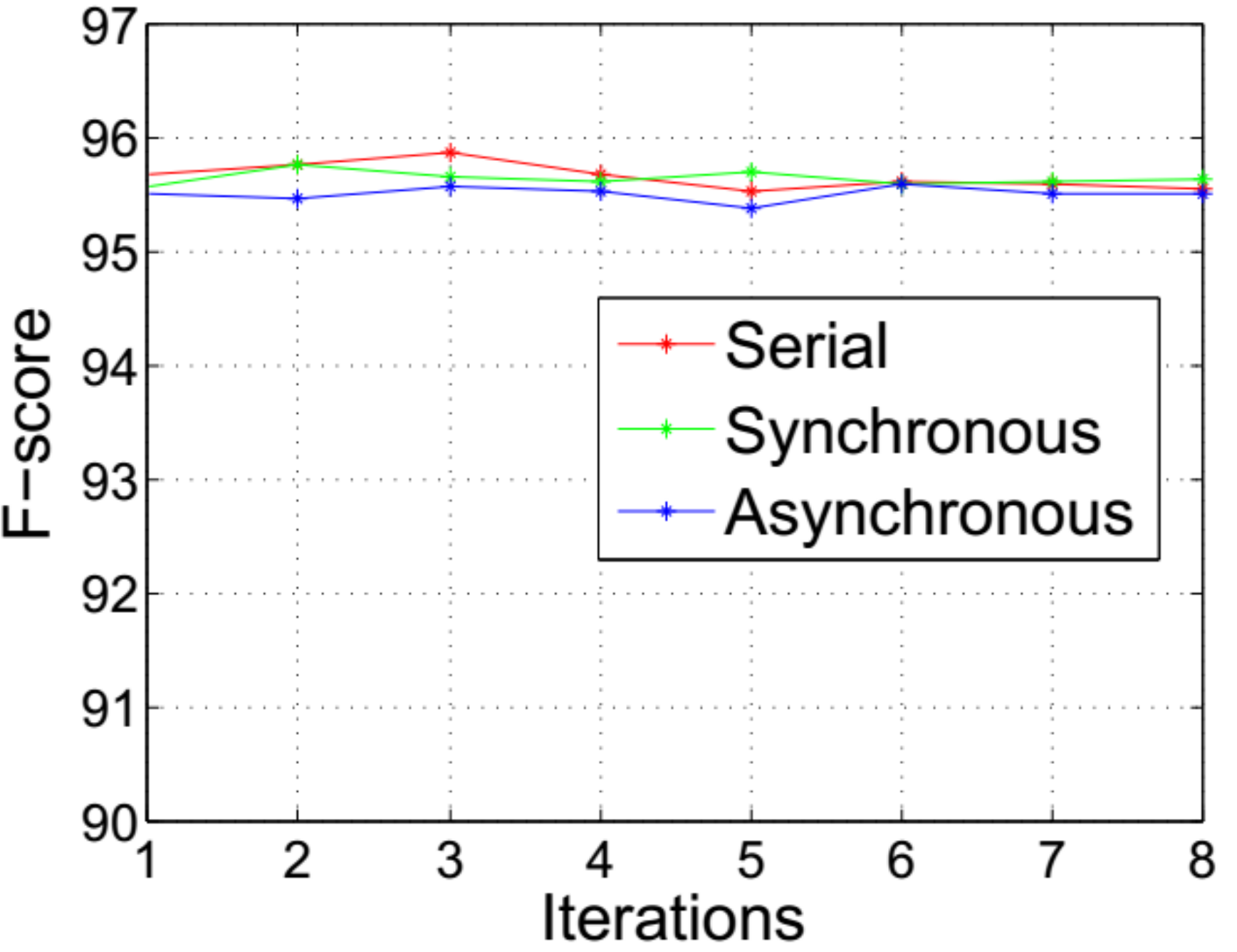}}
\centerline{(b) PKU}
\end{minipage}
\begin{minipage}{0.32\linewidth}
\centerline{\includegraphics[width=4.1cm]{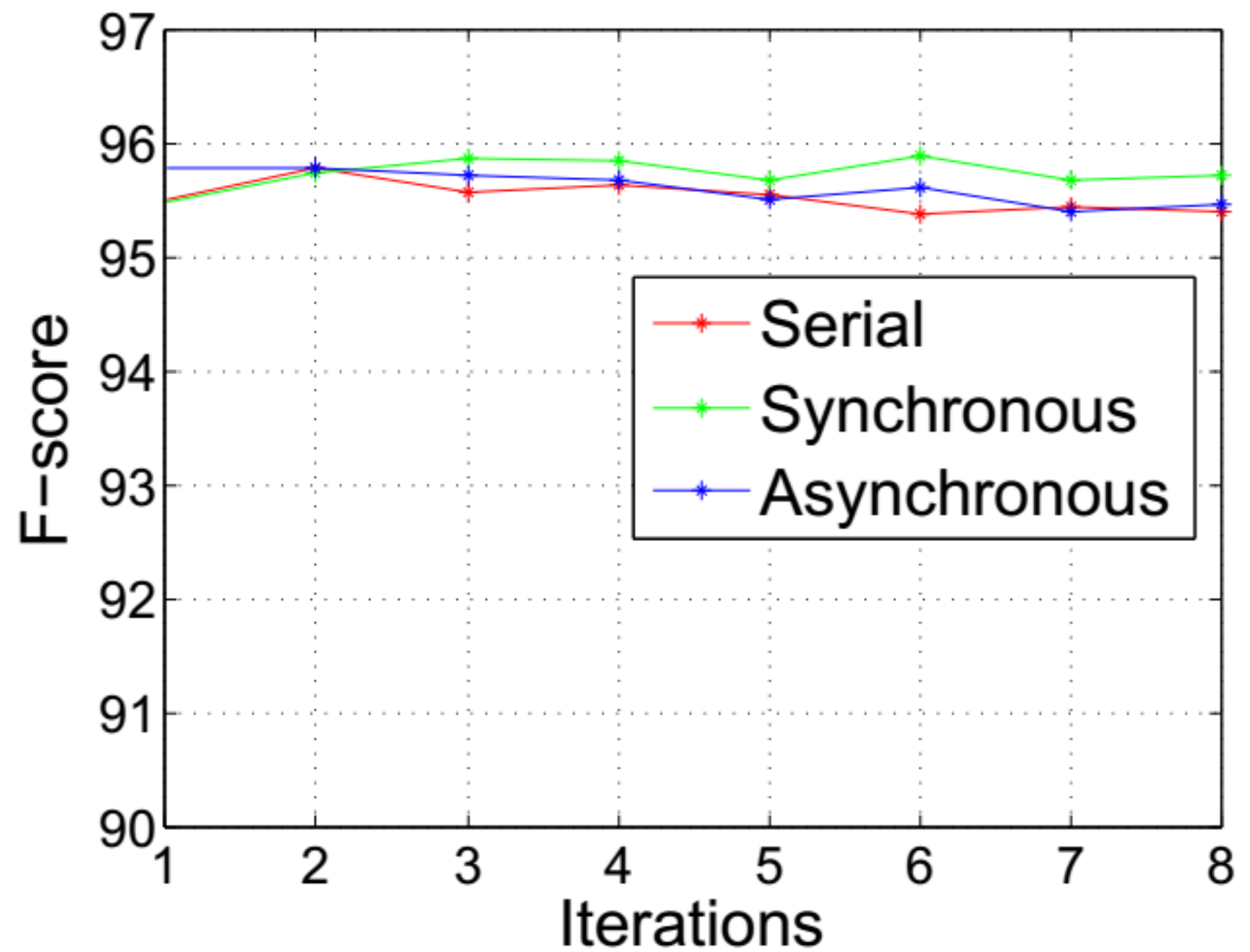}}
\centerline{(c) CTB}
\end{minipage}
\caption{Comparisons of F-score performance among serial, synchronous and asynchronous algorithms on three datasets.}
\label{parallelfscore}
\end{figure}

Table \ref{transferresults} shows the improvement of our proposed approach. The proposed approach is compared with Bi-LSTM which is a competitive and widely used model for neural word segmentation. Experiment results show that our proposed approach achieves substantial improvement on low-resource datasets: 2.3\% and 1.5\% F-score on PKU and CTB datasets. Besides, the error rate is decreased by 34.3\% and 26.3\%.

\textbf{Transfer Learning.} The improvement of transfer learning is shown in Table \ref{transferresults}. We choose MSR as a high-resource dataset. Results on PKU and CTB datasets all show improvement: 1.0\% F-score on PKU dataset and 0.5\% on CTB dataset. A high-resource dataset not only decreases the number of out-of-vocabulary words, but also improves results of in-vocabulary words. The size of PKU dataset is far less than that of CTB dataset and we achieve the better improvement on PKU dataset. It shows that our transfer learning method is more efficient on datasets with lower resource. 

\textbf{Unified Global-Local Neural Networks.} We reconstruct  some of state-of-the-art neural models in this paper: Bi-LSTM and GRNN. Table \ref{tab2} shows that our model outperforms baselines on low-resource datasets: PKU and CTB. It proves that combining several weak features is an effective way to improve the performance on low-resource datasets.

\textbf{Comparisons with State-of-the-art Models.} Table \ref{tabneural} shows comparisons between our work and latest neural models on low-resource datasets: PKU and CTB. Experiment results show that our work largely outperforms state-of-the-art models which are very competitive. Since the dictionary used in Chen et al.~\cite{chen} is not publicly released, our work is not comparable with Chen et al.~\cite{chen}. 

We also compare our work with traditional models on low-resource datasets: PKU and CTB, several of which take advantage of a variety of feature templates and dictionaries. As shown in Table \ref{tab3}, our work achieves state-of-the-art results. Although our model only uses simple bigram features, it outperforms the previous state-of-the-art methods which use more complex features.

%
%

\textbf{Mini-Batch Asynchronous Parallel Learning.} We run the proposed model in asynchronous, synchronous and serial modes to analyze the parallel efficiency. The number of threads used in asynchronous and synchronous modes is 15. The comparisons are shown in Figure \ref{parallelfscoreq} and \ref{parallelfscore}. It can be clearly seen that the asynchronous algorithm achieves the best speedup ratio without decreasing F-score compared with synchronous and serial algorithms. The asynchronous parallel algorithm is almost 5x faster than the serial algorithm.

\section{Conclusions}

The major problem of low-resource word segmentation is insufficient training data. Thus, we propose a transfer learning method to improve the task by leveraging high-resource datasets. Experiment results show that our work largely improves the performance on low-resource datasets compared with state-of-the-art models. Finally, our parallel training method brings substantial speedup and the training speed is almost 5x faster than a serial mode.

\section{Acknowledgments}
We thank the anonymous reviewers for their valuable comments. This work was supported in part by National High Technology Research and Development Program of China (863 Program, No. 2015AA015404), National Natural Science Foundation of China (No. 61673028).

%
%

\end{document}